\documentclass[10pt,twocolumn,letterpaper]{article}
\usepackage[pagenumbers]{cvpr}

\usepackage{cuted}
\usepackage{currfile}
\usepackage{caption}

\usepackage{amsmath}
\usepackage{amssymb}
\usepackage{booktabs}
\usepackage{float}
\usepackage{flafter}
\usepackage{array}
\usepackage{tabularx}
\usepackage{multirow}
\usepackage{natbib}
\usepackage{caption}
\graphicspath{{./}}
\definecolor{cvprblue}{rgb}{0.21,0.49,0.74}
\usepackage[pagebackref,breaklinks,colorlinks,allcolors=cvprblue]{hyperref}
\def\confName{CVPR}
\def\confYear{2027}
\newcommand{\HSG}{HSG}
\newcommand{\HSGNav}{HSGNav}
\newcommand{\aaaitabsetup}{\footnotesize\setlength{\tabcolsep}{2pt}}
\title{\HSGNav: Hypothesis Scene Graphs for Multi-Scale Lifelong Navigation under Object Relocation}
\author{%
Kaiyao Liao\textsuperscript{1}\quad
Qingfeng Li\textsuperscript{1}\quad
Chen Chen\textsuperscript{2}\quad
Qing Sun\textsuperscript{2}\quad
Ningbo Gu\textsuperscript{2}\\[3pt]
Xiaoze Wu\textsuperscript{1}\quad
Haochen Zhao\textsuperscript{1}\quad
Ying Li\textsuperscript{1}\quad
Jianwei Niu\textsuperscript{1,*}\\[6pt]
{\normalsize\textsuperscript{1}Beihang University, Beijing, China}\\
{\normalsize\textsuperscript{2}Hangzhou Innovation Institute, Beihang University, Hangzhou, China}\\[3pt]
{\small\textsuperscript{*}Corresponding author.}%
}
\hypersetup{%
  pdftitle={HSGNav: Hypothesis Scene Graphs for Multi-Scale Lifelong Navigation under Object Relocation},
  pdfauthor={Kaiyao Liao, Qingfeng Li, Chen Chen, Qing Sun, Ningbo Gu, Xiaoze Wu, Haochen Zhao, Ying Li, Jianwei Niu},
  pdfsubject={Multi-scale lifelong object navigation under object relocation},
  pdfkeywords={HSGNav, hypothesis scene graphs, lifelong navigation, object relocation}%
}
\begin{document}
\flushbottom
\maketitle
\begin{strip}
\centering
\includegraphics[width=\textwidth]{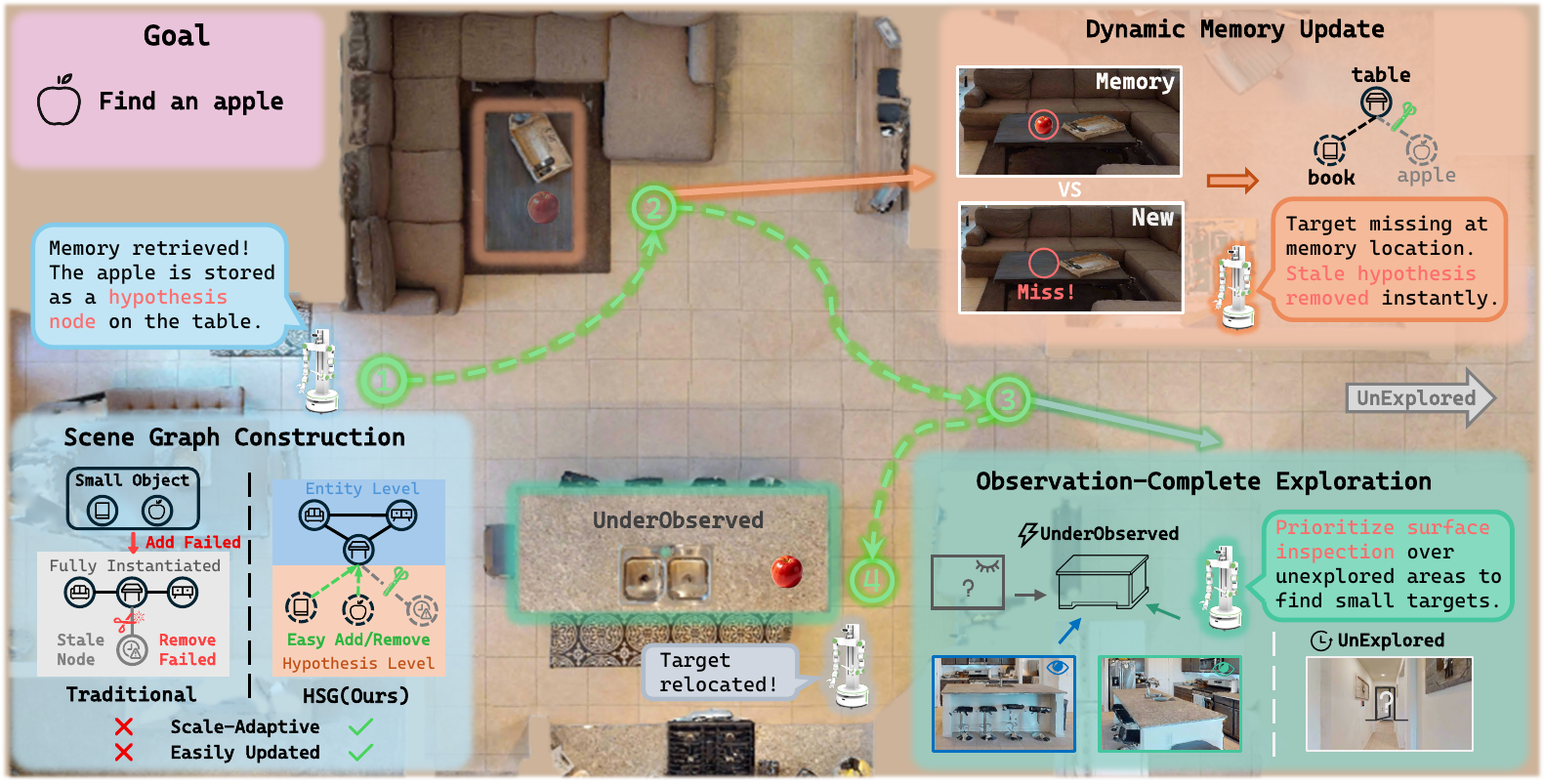}
\captionsetup{hypcap=false}
\captionof{figure}{\textbf{\HSGNav{} searches across scales and revises memory under object relocation.} Our approach tackles surface-object relocation and multi-scale targets through three core mechanisms. \textbf{(Left)} Unlike flat object graphs, our Hypothesis Scene Graph decouples stable furniture (entities) from revisable surface items (hypotheses), enabling local revision of stale memory. \textbf{(Top Right)} When the agent visits a remembered location and finds the target missing, Belief-Driven Verification locally revises the contradicted hypothesis without rebuilding the 3D map. \textbf{(Bottom Right)} Upon memory failure, Observation-Complete Exploration can prioritize inspecting under-observed local surfaces over distant unexplored frontiers to rediscover the relocated target.}
\captionsetup{hypcap=true}
\label{fig:teaser}
\end{strip}

\begin{abstract}
In real homes, assistive robots must repeatedly locate targets from large furniture to tiny tabletop items, as people move objects between household surfaces.
Uniformly instantiating detections at different scales as 3D objects can be brittle for small on-surface items; without explicit object-state validation, stale entries may persist after scene changes.
Our key insight is to separate persistent 3D furniture entities from revisable surface-object hypotheses, retaining their visual evidence without requiring independent object geometry.
We formalize this design as the Hypothesis Scene Graph (\HSG), which preserves small-object evidence and supports local revision without rebuilding the stable entity topology.
Building on \HSG{}, \textbf{\HSGNav} uses this separation for multi-scale lifelong navigation: retrieval selects entity or hypothesis evidence according to target scale, exploration acquires additional surface observations, and verification confirms target presence or invalidates contradicted hypothesis records.
Because existing lifelong benchmarks such as GOAT-Bench rarely require mid-episode memory invalidation or multi-scale search, we introduce \textbf{FAROS-Bench} (\textit{Find And Re-find Objects across Scales}) for mid-episode relocation of surface objects and multi-scale find goals.
Experiments show that \HSGNav{} performs competitively on GOAT-Bench while substantially outperforming prior methods on FAROS-Bench.
\end{abstract}

\section{Introduction}

As embodied agents move from one-shot demos to persistent household assistance, lifelong object navigation~\cite{goatbench} becomes central.
Rather than resetting the map for every query, an agent must continuously find open-vocabulary targets\cite{hm3d-ovon} in the same environment while reusing and updating scene memory under real-world dynamics.
This setting is essential for household robotics and assistive agents, and it is harder than single-episode object navigation\cite{cows,esc,openfmnav,vlfm}: observations remain incomplete, targets span heterogeneous scales, and memories can become stale when the scene changes.

Many open-vocabulary navigation systems ground search in 3D scene graphs built from RGB-D observations~\cite{conceptgraphs,hovsg,sgnav,unigoal,3dmem,msgnav,cmar}.
These representations support object-level retrieval, but small on-surface objects often yield intermittent detections and unstable depth, making consistent cross-view association difficult.
Their observations may therefore fail to accumulate sufficient support for a distinct 3D instance, leaving valid object evidence filtered out or incorrectly merged with surrounding furniture.
Even when they are instantiated, object nodes may be retained without explicit state validation or removal, allowing stale entries to bias subsequent navigation after an object is relocated or disappears.

Our key insight, embodied in the Hypothesis Scene Graph (\HSG), is to separate persistent 3D furniture entities from revisable surface-object hypotheses that require no independent 3D geometry.
Furniture entities provide spatial anchors, while hypotheses retain visual evidence of objects on their surfaces.
This separation allows small-object evidence to enter memory without reliable instance geometry and to be locally revised after relocation, while preserving the stable entity topology.
Building on \HSG{}, \textbf{\HSGNav} uses the two layers for multi-scale search: retrieval selects entity or hypothesis evidence according to target scale, exploration inspects under-observed surfaces or unseen areas, and verification confirms target presence or invalidates contradicted hypothesis records.
Together, these operations make hypothesis evidence available for navigation while allowing its validity to change independently of the persistent furniture geometry.

Existing lifelong benchmarks such as GOAT-Bench~\cite{goatbench} mainly evaluate static, furniture-dominated goals and rarely require mid-episode memory invalidation or multi-scale search within the same episode.
We therefore introduce \textbf{FAROS-Bench} for the \textit{Find And Re-find Objects across Scales} (FAROS) task, evaluating sequential search from tiny surface objects to large furniture, with mid-episode surface-object relocation under stable furniture anchors.
Experiments show that \HSGNav{} performs competitively on GOAT-Bench while substantially outperforming prior methods on FAROS-Bench.

Our contributions are summarized as follows:
\begin{itemize}
    \item We propose \textbf{\HSG}, separating persistent 3D furniture entities from revisable surface-object hypotheses without independent object geometry.
    \item Building on \HSG{}, we develop \textbf{\HSGNav}, which retrieves, enriches, and verifies hypothesis evidence for multi-scale lifelong search, locally revising surface memory while preserving the stable entity topology.
    \item We introduce \textbf{FAROS-Bench} for multi-scale lifelong object navigation with mid-episode relocation of surface objects, and demonstrate strong results on both GOAT-Bench and FAROS-Bench.
\end{itemize}

\section{Related Work}

\subsection{3D Scene Graphs and Embodied Memory}

Dense 3D representations ground Vision-Language Foundation Model features~\cite{clip,blip,seem} for open-vocabulary queries~\cite{openscene,CLIP-Fields,Open-fusion,CLIP-FO3D,3D-Concept-Grounding}, but are costly to maintain and update.
Explicit 3D scene graphs instead use compact object--relation topologies~\cite{hydra,Kimera,SceneGraphFusion}, although traditional variants encode relations as restrictive text and discard visual nuance~\cite{3D-Scene-Graph,VGfM,conceptgraphs,3D-Semantic-Scene-Graph}.
Recent methods retain richer evidence through multi-view snapshots (3D-Mem~\cite{3dmem}) or image-patch edges (MSGNav~\cite{msgnav}).
Contemporaneous OASIS~\cite{liao2026oasis} organizes mapped object relations and representative views around stable semantic anchors.
However, these methods rely on instance-level 3D construction for both furniture and small objects.
\HSG{} instead retains surface-object evidence as revisable hypotheses without requiring independent 3D geometry, while preserving persistent furniture anchors.

\subsection{Lifelong Navigation under Object Relocation}

Recent map-based approaches also investigate scene-graph maintenance and object relocation~\cite{dualmap,idu,openin}.
DualMap updates its abstract map online and reselects navigation candidates when a target is not found at its expected location.
OpenIN initializes a carrier--object graph from pre-collected RGB-D data and updates carried-object states during navigation.
OASIS~\cite{liao2026oasis} verifies targets at remembered viewpoints and combines transient visual observations with semantic reasoning to re-find moved objects.
Contemporaneous IVAM-Nav~\cite{huang2026lisonav} uses multi-view surface inspection and viewpoint-anchored memory to revisit and revalidate stale information.
\HSGNav{} separates persistent 3D furniture entities from revisable surface-object hypotheses without independent object geometry.
LiSoNav-Eval moves objects between episodes; FAROS-Bench re-queries the same target after two relocations within an episode and includes furniture-scale goals.
Target-conditioned verification locally invalidates contradicted hypothesis descriptions, preserving the persistent entity topology and excluding rejected evidence from subsequent retrieval.

\section{Method}

\subsection{Overview}

\begin{figure*}[t]
    \centering
    \includegraphics[width=\textwidth]{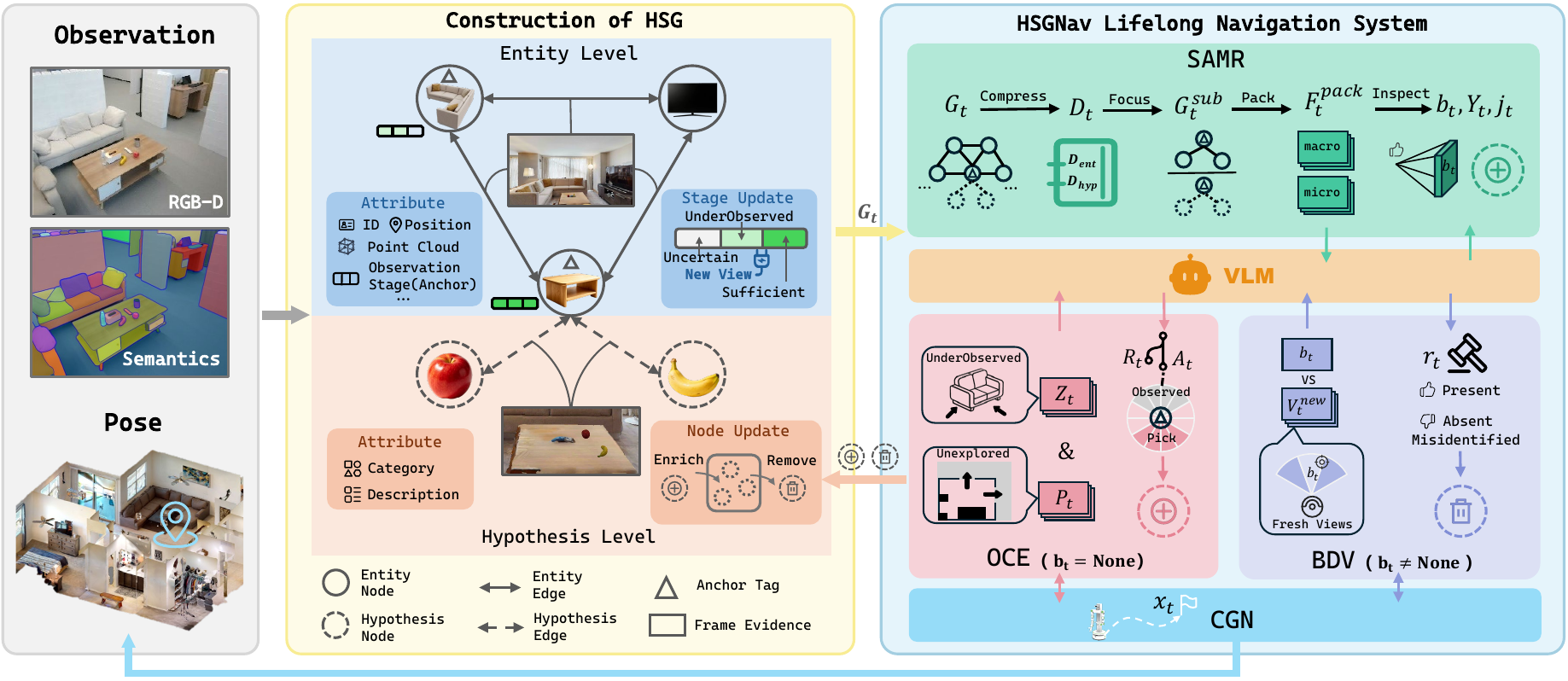}
\caption{\textbf{Overview of \HSGNav.}
Streaming RGB-D observations build a two-level \HSG{}: persistent 3D entities serve as anchors, while on-surface objects are represented as revisable hypotheses linked to them. SAMR retrieves scale-appropriate visual evidence for the target. When retrieval fails, OCE selects a new observation goal; CGN executes the selected goal, and BDV verifies retrieved targets using fresh views. New observations and SAMR add or enrich hypotheses, while BDV invalidates contradicted descriptions and removes hypotheses with no valid descriptions remaining. Dashed plus and trash icons denote hypothesis updates propagated to the \HSG{}.}
\label{fig:pipeline}
\end{figure*}

\HSGNav{} separates persistent 3D furniture entities from revisable surface-object hypotheses, retaining small-object evidence without independent object geometry (Sec.~\ref{sec:hsg}).
Scale-adaptive memory retrieval (SAMR) selects entity or hypothesis evidence according to target scale (Fig.~\ref{fig:pipeline}).
A hit triggers committed goal navigation (CGN) and belief-driven verification (BDV) with fresh views.
BDV invalidates contradicted description records, removing hypotheses only when no valid records remain while preserving the stable entity topology (Sec.~\ref{sec:vdda}).
When retrieval fails, observation-complete exploration (OCE) selects frontiers or under-observed surfaces for inspection, also executed by CGN.
New observations and SAMR enrich hypotheses, enabling evidence acquisition and revision.

\subsection{\HSG{} Structure and Incremental Construction}
\label{sec:hsg}

\HSG{} is a hybrid property graph $G_t=(V_t,E_t)$, with $V_t=V_{\mathrm{ent}}\cup V_{\mathrm{hyp}}$ and $E_t=E_{\mathrm{ent}}\cup E_{\mathrm{hyp}}$.
Rather than becoming nodes, RGB-D frames remain edge evidence $e=(u,v,F_e)$, while VLM-generated descriptions are stored as attributes of hypothesis nodes, enabling local hypothesis maintenance without rebuilding the stable entity topology.
Each frame $I_t$ at pose $\xi_t$ updates the nodes and then the edges of $G_{t-1}$.

\paragraph{Entity nodes and edges.}
Entity nodes $V_{\mathrm{ent}}$ represent observed 3D objects with semantic and geometric attributes.
Stable furniture and support surfaces form semantic anchors $V_{\mathrm{anc}}\subseteq V_{\mathrm{ent}}$, with $A_a=\langle O(a),\eta_t(a)\rangle$: $O(a)$ accumulates RGB-D views, and $\eta_t(a)\in\{\textsc{Uncertain},\textsc{UnderObserved},\textsc{Sufficient}\}$ summarizes coverage for OCE (Sec.~\ref{sec:afap}).
Entity edges $e_{ij}^{\mathrm{ent}}=(v_i^{\mathrm{ent}},v_j^{\mathrm{ent}},F_{ij})$ connect co-visible pairs and store their witnessing frames.

Following ConceptGraph~\cite{conceptgraphs}, each view $I_t$ is back-projected and matched to an entity by spatial and visual similarity; matched views update node attributes, while unmatched views spawn nodes.
For an anchor $a$, the association also appends $I_t$ to $O(a)$ and updates $e_{ij}^{\mathrm{ent}}$ with $I_t\in F_{ij}$ for every co-observed entity pair.
For each anchor, we select a top-$K$ subset of $O(a)$ to maximize an inspection score that determines its observation state $\eta_t(a)$.
\begin{align}
Q(f) &= \tfrac{1}{3}\bigl(A(f)+E(f)+D(f)\bigr),
\label{eq:micro-q}\\
T(a) &=
\max_{\substack{U\subseteq O(a)\\ |U|\le K}}
\Bigl(\tfrac{1}{K}\sum_{f\in U} Q(f) + B(U)\Bigr),
\label{eq:inspection-total}
\end{align}
where $A(f),E(f)\in[0,1]$ score mask area and completeness against image-boundary truncation, respectively, and $D(f){=}\max(0,\,1{-}d_f/d_{\mathrm{ref}})$ scores camera-to-anchor proximity ($K{=}5$, $d_{\mathrm{ref}}{=}3$\,m; empty slots contribute zero when $|U|{<}K$).
$U_t(a)\subseteq O(a)$ denotes a maximizing top-$K$ set for $T(a)$.
We divide the viewing directions around $a$ into five equal $72^\circ$ sectors. We set $B(U){=}0.1$ if the views in $U$ come from at least three sectors, and $B(U){=}0$ otherwise.
With $n_{\mathrm{obs}}^{\min}{=}2$, $\tau{=}0.4$, and anchor-category consistency check $\iota_t(a)$, the observation stage is
\begin{equation}
\eta_t(a)=
\begin{cases}
\textsc{Uncertain}, & n_{\mathrm{obs}}(a){<}n_{\mathrm{obs}}^{\min}\ \vee\ \neg\iota_t(a), \\[2pt]
\textsc{UnderObserved}, & \iota_t(a) \wedge T(a){<}\tau, \\[2pt]
\textsc{Sufficient}, & \iota_t(a) \wedge T(a){\ge}\tau,
\end{cases}
\label{eq:anchor-stage}
\end{equation}
where $n_{\mathrm{obs}}(a)=|O(a)|$ and $\iota_t(a)$ holds when at least three views in $O(a)$ carry the anchor category $C_a$.
Anchors marked \textsc{UnderObserved} are prioritized for close-range re-observation (Sec.~\ref{sec:nav}).

\paragraph{Hypothesis nodes and edges.}
Hypothesis nodes $V_{\mathrm{hyp}}$ represent on-surface object classes that are not promoted to independent 3D entities.
Each node $v_{ak}^{\mathrm{hyp}}=\langle c_k,\mathcal{D}_{ak}\rangle$ represents one hypothesized class $c_k\in\Sigma$ on anchor $a$: an anchor has at most one node per class, but may link to nodes for many classes. The node attribute $\mathcal{D}_{ak}$ stores anchor-local appearance records, each with a description ID and one or more supporting views.
The existence of $v_{ak}^{\mathrm{hyp}}$ denotes a supported surface hypothesis whose visual evidence is maintained separately from entity geometry and locally revised when contradicted by observations. Hypothesis edges $e_{ak}^{\mathrm{hyp}}=(v_a^{\mathrm{ent}}, v_{ak}^{\mathrm{hyp}}, F_{ak})$ link anchor $a$ to $v_{ak}^{\mathrm{hyp}}$ and store its supporting surface views $F_{ak}$.

For each observed anchor, we apply the YOLO-World~\cite{yolo} detector used for entity construction to an expanded image crop at confidence $0.1$. A larger upper margin includes objects on the support surface; crop settings are provided in the supplementary material.
Each detected non-anchor class $c_k$ creates a hypothesis node $v_{ak}^{\mathrm{hyp}}$ on $a$ if no node for $c_k$ already exists; its hypothesis edge is initialized with the supporting surface crop from $I_t$. Subsequent supporting crops are appended to the existing hypothesis edge without creating duplicate nodes. Description records are added or enriched only when SAMR's VLM jointly inspects the packed views.
Hypothesis nodes are never fused into the 3D map: streamed surface observations and SAMR inspection can add hypotheses, while BDV locally invalidates contradicted description records during navigation (Sec.~\ref{sec:nav}).

\subsection{\HSGNav{} Lifelong Navigation System}
\label{sec:nav}

The complete control-loop pseudocode is provided in the supplementary material as Algorithm~S1.
When SAMR returns a best-view index, the agent commits its frame pose as a \textsc{Verify} destination $x_t$, carries the matched description record for a micro hit, and navigates via CGN before running BDV; otherwise OCE proposes an \textsc{Inspect} or \textsc{Frontier} destination, whose new observations may create surface hypotheses before the next retrieval step.

\subsubsection{Scale-Adaptive Memory Retrieval (SAMR)}
\label{sec:samr}

As $G_t$ grows, only a small target-relevant slice is useful per subtask.
SAMR adapts this slice to the target scale through four steps: compress, focus, pack, and inspect.

\paragraph{Compress.}
We compress $G_t$ into a text-only index
\begin{equation}
D_t = \bigl\{(a,\, D_{\mathrm{ent}}(a),\, D_{\mathrm{hyp}}(a))\bigr\}_{a\in\hat{V}_t^{\mathrm{anc}}},
\label{eq:ret-index}
\end{equation}
where $\hat{V}_t^{\mathrm{anc}}$ lists semantic anchors $a\in V_{\mathrm{anc}}$ plus one virtual anchor $a^{\mathrm{unrel}}$ for observations not tied to any semantic anchor.
For each anchor, $D_{\mathrm{ent}}(a)$ summarizes co-visible entity classes from $e_{ij}^{\mathrm{ent}}$, while $D_{\mathrm{hyp}}(a)$ summarizes surface-object classes and their stored descriptions. The virtual anchor $a^{\mathrm{unrel}}$ has only an entity summary.
No frames are loaded at this stage.

\paragraph{Focus.}
Given target $y$---passed to the VLM in its native form (category name, language description, or image)---the VLM selects a compact subgraph from the index:
\begin{equation}
G_t^{\mathrm{sub}} = \bigl(V_t^{\mathrm{sub}}, E_t^{\mathrm{sub}}\bigr)
\leftarrow
\mathrm{VLM}(D_t,y).
\label{eq:anchor-select}
\end{equation}
The VLM infers the target scale and selects hypothesis summaries $D_{\mathrm{hyp}}(a)$ for small or on-surface targets and entity summaries $D_{\mathrm{ent}}(a)$ for furniture-scale targets. These selections determine the nodes $V_t^{\mathrm{sub}}$ and edges $E_t^{\mathrm{sub}}$ of $G_t^{\mathrm{sub}}$. At most $K_{\mathrm{anc}}{=}3$ anchor-linked node--edge groups are retained, with entity and hypothesis groups attached to the same anchor counted separately.

\paragraph{Pack.}
Pack loads the visual evidence stored on the selected edges $E_t^{\mathrm{sub}}$ into $F_t^{\mathrm{pack}}$. Entity edges provide co-visibility frames as macro views, while hypothesis edges provide surface-supporting frames as micro views. Conditioned on the inferred target scale, Pack prioritizes micro views for small or on-surface targets and macro views for furniture-scale targets; surface views are prioritized by each anchor's top-$K$ set $U_t(a)$ (Eq.~\eqref{eq:inspection-total}).

\paragraph{Inspect.}
The packed frames $F_t^{\mathrm{pack}}=\{f_0,\ldots,f_{n-1}\}$ are jointly sent to the VLM:
\begin{equation}
\left(Y_t,\, b_t,\, j_t\right)
\leftarrow
\mathrm{VLM}\!\left(F_t^{\mathrm{pack}},y\right),
\label{eq:memory-inspect}
\end{equation}
where $\mathcal{R}_\ell$ lists category--appearance pairs $(c,d)$ for visible objects in $f_\ell$, and $Y_t=\{(\ell,\mathcal{R}_\ell)\}_{\ell=0}^{n-1}$. Separate instances may yield identical pairs. The best-view index is $b_t\in\{0,\ldots,n{-}1\}\cup\{\textsc{None}\}$, with \textsc{None} indicating no hit.
For a micro hit, $j_t$ selects the target-matched entry in $\mathcal{R}_{b_t}$ using the query category and any appearance description or reference image; otherwise, $j_t{=}\textsc{None}$.
Category goals match by class; description and image goals require evidence for the specified instance.
For anchor $a$ and class $c_k$, $\mathcal{I}_{ak}(Y_t)=\{(\ell,j,d)\mid f_\ell\in F_a^{\mathrm{micro}},\ (c_k,d)=\mathcal{R}_\ell[j]\}$ collects entries from the anchor's packed micro views. SAMR finds or creates the unique hypothesis node $v_{ak}^{\mathrm{hyp}}$ and merges these entries with its appearance records:
\begin{equation}
\mathcal{D}_{ak} \leftarrow
\mathrm{Merge}\!\left(\mathcal{D}_{ak},\mathcal{I}_{ak}(Y_t)\right),
\label{eq:hypothesis-merge}
\end{equation}
where Merge associates observations with one description ID only when visual and spatial context unambiguously establish correspondence to the same object at the same anchor-relative location, retaining their source frames. Otherwise, observations remain separate records even when their categories and appearance descriptions coincide; these anchor-local records retain appearance evidence and supporting views for subsequent target-conditioned retrieval and verification. The selected entry $(b_t,j_t)$ resolves to a merged record $\delta_t$, which BDV can invalidate while retaining other same-class records when fresh views distinguish the contradicted candidate.

When retrieval and OCE yield no candidates, SAMR selects anchors from the text index, including sufficiently observed ones, for BDV without record invalidation, repeating until success or budget exhaustion.

\subsubsection{Observation-Complete Exploration (OCE)}
\label{sec:afap}

Geometric exploration does not guarantee sufficient surface observation, so small targets may remain unseen in mapped areas.
When SAMR returns $b_t{=}\textsc{None}$, the agent must decide whether to explore unseen space or revisit a known under-observed anchor.
Let $A_t=\{a:\eta_t(a){=}\textsc{UnderObserved}\}$ and $R_t$ denote under-observed anchors and candidate frontiers, with preview images $Z_t$ and $P_t$.
A single VLM call selects the next destination:
\begin{equation}
u_t \leftarrow \mathrm{VLM}(Z_t,P_t,y),
\label{eq:afap}
\end{equation}
where $u_t$ is either a frontier $r\in R_t$ or an inspect goal for an anchor $a\in A_t$.
For a frontier, OCE follows standard frontier-based exploration~\cite{3dmem} to its center; frontiers are free--unexplored boundaries of the TSDF occupancy grid that expand coverage into unmapped space.
For an under-observed anchor $a^*$, each view in $O(a^*)$ marks its previously defined viewing sector as covered.
OCE selects the uncovered sector whose center direction is farthest from those already covered, then places the agent on the standoff circle ($d_{\mathrm{obs}}{=}1$\,m) such that it faces $a^*$ from that sector:
\begin{align}
\Omega(a^{*}) &= \bigl\{p\mid \|p-P_{a^{*}}\|=d_{\mathrm{obs}}\bigr\},
\label{eq:obs-circle}\\
x_{t} &= \arg\max_{p\in\Omega(a^{*})}
  \mathrm{gap}\bigl(\mathrm{bin}(p),\,O(a^{*})\bigr).
\label{eq:obs-viewpoint}
\end{align}
Here $P_{a^*}$ is the anchor center, $\Omega(a^*)$ is the circle at inspection distance $d_{\mathrm{obs}}$, $\mathrm{bin}(p)$ identifies the viewing sector associated with position $p$, and $\mathrm{gap}(\cdot)$ measures its minimum angular distance from the sectors already covered by $O(a^*)$.
The selected waypoint becomes the next \textsc{Inspect} destination.
Surface observations gathered during inspection may create new hypothesis nodes on the observed anchor.

\subsubsection{Belief-Driven Verification (BDV)}
\label{sec:vdda}

A SAMR hit may be ambiguous or outdated, so when $b_t\neq\textsc{None}$, the agent revisits the selected frame pose $\xi(f_{b_t})$ for verification.
Let $\tau(b_t)\in\{\mathrm{macro},\mathrm{micro}\}$ indicate whether $f_{b_t}$ comes from an entity edge or an anchor's surface set $U_t(a)$.
Macro verification captures a three-view ring at the replay pose ($\pm60^\circ$); micro verification additionally approaches the linked anchor to $d_{\mathrm{obs}}$ and captures a close-range ring.
BDV compares the resulting views $V_t^{\mathrm{new}}$ with the saved frame and target query, including the selected description $d_{\delta_t}$ for a micro hit (empty for a macro hit):
\begin{equation}
r_t \leftarrow
\mathrm{VLM}\!\left(f_{b_t},\,V_t^{\mathrm{new}},\,y,\,d_{\delta_t}\right),
\label{eq:arrival-verify}
\end{equation}
where $r_t\in\{\textsc{Present},\textsc{Misidentified},\textsc{Absent}\}$ distinguishes a confirmed target from a retrieval error or a vacated location.
A \textsc{Present} verdict terminates navigation. For a rejected micro hit, BDV invalidates the selected record $\delta_t$ only when fresh views contradict the record itself. Valid records that merely mismatch the query are excluded only for the current subtask. Invalidation retains the frame and all other records:
\begin{equation}
\thickmuskip=3mu
\mathcal{N}_{ak}\leftarrow\mathcal{N}_{ak}\cup\{\delta_t\},\;
v_{ak}^{\mathrm{hyp}}\text{ removed}\iff\mathcal{D}_{ak}\subseteq\mathcal{N}_{ak},
\label{eq:bdv-local-update}
\end{equation}
where $\mathcal{N}_{ak}$ stores invalidated records keyed by their description IDs. Withdrawal excludes all cached entries linked to that ID from retrieval without regenerating them from old views; fresh observations may supply a new record. Thus, BDV revises appearance evidence locally while retaining shared frames.

\begin{figure}[t]
\centering
\includegraphics[width=\linewidth]{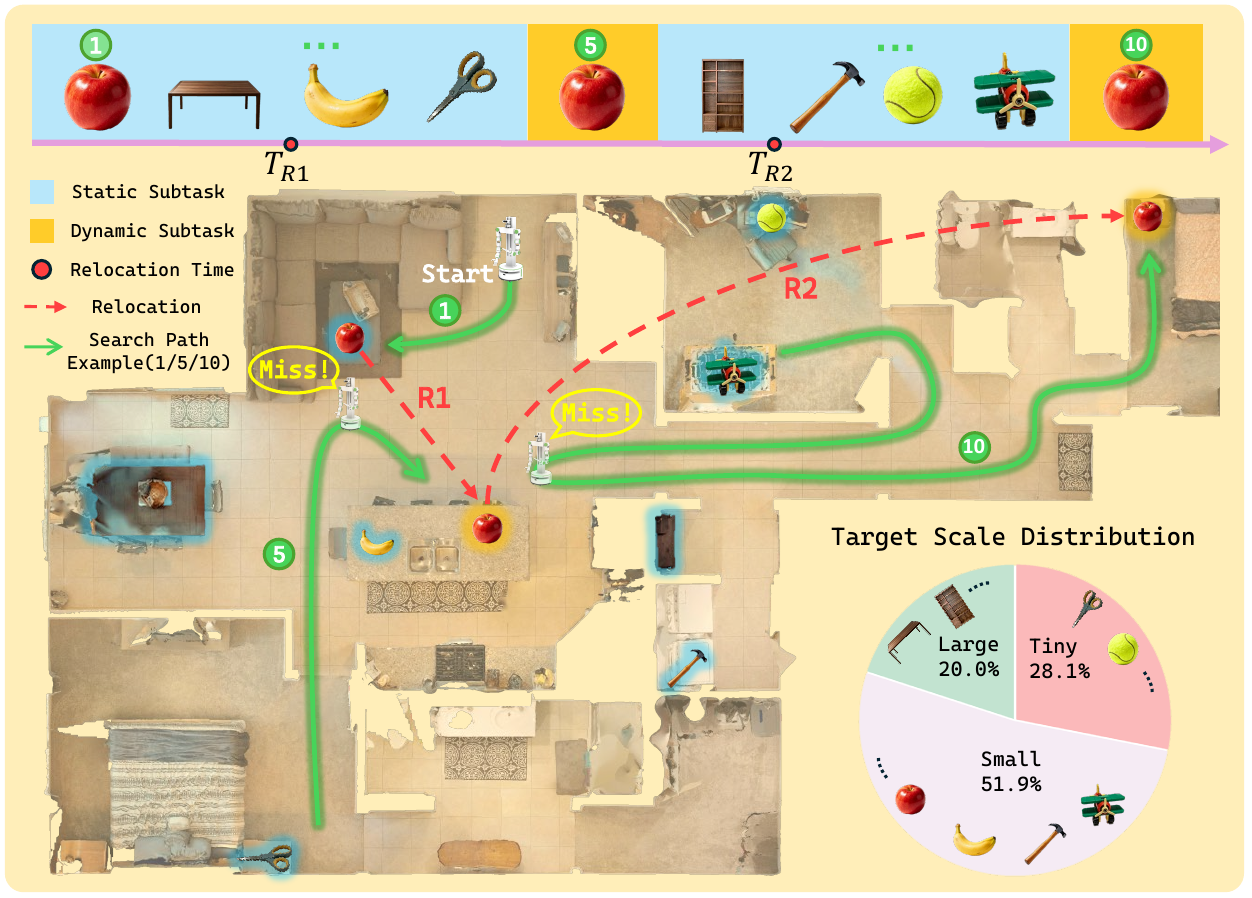}
\caption{\textbf{Overview of FAROS-Bench.} Each episode interleaves static and dynamic find subtasks; the same movable target is queried at steps~1, 5, and~10 with two intervening relocations, while goals span tiny, small, and large scales.}
\label{fig:dynamic_bench}
\end{figure}

\subsubsection{Committed Goal Navigation (CGN)}
\label{sec:cepi}

CGN commits to the waypoint $x_t$ selected by SAMR or OCE and navigates until arrival without intermediate VLM replanning, reducing repeated queries and goal switching~\cite{3dmem,msgnav}.
The high-level destination stays fixed while perception and incremental graph construction continue.
During transit, the agent captures a ring of views every $\Delta d{=}1$\,m~\cite{3dmem} and incorporates them into \HSG{} (Sec.~\ref{sec:hsg}).
Upon arrival, \textsc{Verify} goals trigger BDV, whereas \textsc{Inspect} and \textsc{Frontier} goals return control to SAMR.

\section{Experiments}
\label{sec:experiments}

\subsection{Experimental Setup}

\subsubsection{Datasets}
\label{sec:datasets}
We evaluate lifelong navigation on \textbf{GOAT-Bench}~\cite{goatbench}, with object, description, and image goals in static scenes.
Following prior work~\cite{3dmem}, GOAT-Bench evaluation uses one exploration episode per scene from the \texttt{val\_unseen} split, totaling 278 subtasks across 36 scenes.

\begin{table}[t]
\centering
\caption{\textbf{GOAT-Bench results.} SR and SPL are reported in \%.}
\label{tab:goat_bench}
\footnotesize
\setlength{\tabcolsep}{2pt}
\begin{tabular*}{\linewidth}{@{}@{\extracolsep{\fill}}lccc@{}}
\toprule
\textbf{Method} & \textbf{Training-free} & \textbf{SR}$\uparrow$ & \textbf{SPL}$\uparrow$ \\
\midrule
SenseAct-NN Mono~\cite{goatbench}   & $\times$     & 12.3 & 6.8  \\
CLIP on Wheels~\cite{goatbench}     & $\checkmark$ & 16.1 & 10.4 \\
Modular GOAT~\cite{goatbench}       & $\checkmark$ & 24.9 & 17.2 \\
SenseAct-NN Skill~\cite{goatbench}  & $\times$     & 29.5 & 11.3 \\
VLMnav~\cite{vlmnav}                & $\checkmark$ & 20.1 & 9.6  \\
DyNaVLM~\cite{dynavlm}              & $\checkmark$ & 25.5 & 10.2 \\
TANGO~\cite{tango}                  & $\checkmark$ & 32.1 & 16.5 \\
MTU3D~\cite{mtu3d}                  & $\times$     & 47.2 & 27.7 \\
MSGNav~\cite{msgnav}                & $\checkmark$ & 60.0 & 37.0 \\
3D-Mem~\cite{3dmem}                 & $\checkmark$ & \underline{69.1} & \underline{48.9} \\
\midrule
\textbf{\HSGNav{} (Ours)}           & $\checkmark$ & \textbf{71.2} & \textbf{50.3} \\
\bottomrule
\end{tabular*}
\end{table}

To complement GOAT-Bench's static, furniture-dominated goals, \textbf{FAROS-Bench} evaluates multi-scale search with mid-episode surface-object relocation under stable furniture anchors, testing both memory reuse and stale-memory revision within continuous episodes (Fig.~\ref{fig:dynamic_bench}).
It contains 210 subtasks across 7 HM3D validation scenes, with three 10-subtask episodes per scene.
Goals are large (20.0\%), small (51.9\%), or tiny (28.1\%).
Tiny objects comprise the lowest tertile of YCB categories by collision-mesh bounding-box volume; remaining YCB categories are small, and scene-native furniture is large.
Each episode places eight YCB objects on support surfaces: six serve as find targets and two as distractors.
Two additional goals query furniture categories already present in the HM3D scene.
The same movable target is queried at steps~1, 5, and~10, with relocations after steps~2 and~6.
Steps~5 and~10 are Dynamic tasks; all others are Static tasks.
Dynamic tasks span 21 YCB categories.
Relocations always change support surfaces, with roughly half crossing rooms.
All object placements are manually verified in Habitat-Sim.
Further details appear in the supplementary material.

Following DualMap~\cite{dualmap}, we evaluate memory updates through \textbf{re-localization} in pre-mapped HM3D scenes with stable furniture anchors.
Single-task tests move targets within (In-anchor) or between (Cross-anchor) anchors, using published DualMap scores and OASIS~\cite{liao2026oasis} re-evaluated with GPT-4o for both its LLM and VLM.
Continuous tests comprise 51 sequences from 17 scene--target pairs across three scenes and three seeds, each including three Cross-anchor moves and a fixed-target recheck, with memory retained throughout.
Methods share full-scene RGB-D observations, camera poses, DualMap's instance-construction pipeline, controllers, budgets, and paired random starts; \HSGNav{} disables unknown-region exploration.
Success requires stopping within 1\,m, allowing three candidate attempts per relocation and one per recheck.

\begin{table*}[t]
\centering
\caption{\textbf{FAROS-Bench results (\%).} +Det adds shared surface-crop detections. Representation controls share OCE, BDV, and CGN; module ablations retain full \HSG{}. SR/SPL are grouped by target scale and static/dynamic queries.}
\label{tab:dynamic_bench_mega}
\aaaitabsetup
\begin{tabular*}{\textwidth}{@{}l@{\extracolsep{\fill}}*{12}{c@{\extracolsep{\fill}}}}
\toprule
& \multicolumn{2}{c}{Overall} & \multicolumn{2}{c}{Tiny} & \multicolumn{2}{c}{Small} & \multicolumn{2}{c}{Large} & \multicolumn{2}{c}{Static} & \multicolumn{2}{c}{Dynamic} \\
\cmidrule(lr){2-3} \cmidrule(lr){4-5} \cmidrule(lr){6-7} \cmidrule(lr){8-9} \cmidrule(lr){10-11} \cmidrule(lr){12-13}
\textbf{Method} & SR $\uparrow$ & SPL $\uparrow$ & SR $\uparrow$ & SPL $\uparrow$ & SR $\uparrow$ & SPL $\uparrow$ & SR $\uparrow$ & SPL $\uparrow$ & SR $\uparrow$ & SPL $\uparrow$ & SR $\uparrow$ & SPL $\uparrow$ \\
\midrule
\multicolumn{13}{l}{System Comparison} \\
MSGNav~\cite{msgnav} & 36.7 & 12.5 & 25.4 & 3.7 & 33.0 & 10.3 & 61.9 & 30.5 & 39.3 & 12.4 & 26.2 & 12.8 \\
3D-Mem~\cite{3dmem} & 35.7 & 16.8 & 23.7 & 4.9 & 30.3 & 15.6 & 66.7 & 36.5 & 39.3 & 17.5 & 21.4 & 14.0 \\
MSGNav + Det & 37.6 & 11.8 & 27.1 & 3.4 & 33.9 & 9.7 & 61.9 & 29.1 & 40.5 & 11.7 & 26.2 & 12.1 \\
3D-Mem + Det & 34.3 & 15.1 & 20.3 & 3.8 & 29.4 & 13.5 & 66.7 & 35.1 & 37.5 & 15.7 & 21.4 & 12.8 \\
\midrule
\multicolumn{13}{l}{Representation and Update Controls} \\
Traditional~\cite{conceptgraphs} & 36.2 & 17.0 & 25.4 & 7.1 & 31.2 & 15.5 & 64.3 & 34.8 & 41.1 & 18.0 & 16.7 & 13.0 \\
Multi-modal~\cite{msgnav} & 39.5 & 15.5 & 28.8 & 6.4 & 35.8 & 13.8 & 64.3 & 32.8 & 41.7 & 15.4 & 31.0 & 16.0 \\
Anchor-Centric~\cite{liao2026oasis} & 38.1 & 20.2 & 27.1 & 9.2 & 32.1 & 19.5 & 69.0 & 37.5 & 39.3 & 21.5 & 33.3 & 15.0 \\
Entity-Level Only & 37.1 & 19.8 & 25.4 & 8.3 & 31.2 & 19.2 & 69.0 & 37.4 & 38.1 & 21.0 & 33.3 & 14.8 \\
Full Inst. & 41.0 & 21.4 & 33.9 & 10.5 & 34.9 & 21.5 & 66.7 & 36.4 & 43.5 & 23.5 & 31.0 & 13.2 \\
Full Inst. + Removal & 41.4 & 21.5 & 33.9 & 10.5 & 35.8 & 21.8 & 66.7 & 36.4 & 43.5 & 23.5 & 33.3 & 13.7 \\
\midrule
\multicolumn{13}{l}{Module Ablations} \\
w/o SAMR & 52.4 & 23.9 & 40.7 & 14.5 & 52.3 & 23.8 & 69.0 & 37.3 & 53.0 & 25.1 & 50.0 & 19.1 \\
w/o OCE & 48.1 & 23.7 & 37.3 & 14.0 & 45.9 & 23.4 & 69.0 & 38.1 & 48.2 & 24.5 & 47.6 & 20.5 \\
w/o CGN & \underline{57.1} & 25.4 & \underline{45.8} & \underline{18.2} & \underline{56.9} & 24.3 & \textbf{73.8} & 38.5 & \underline{57.7} & 24.9 & \underline{54.8} & \underline{27.4} \\
w/o BDV & 51.4 & 24.4 & 39.0 & 15.5 & 50.5 & 23.5 & \underline{71.4} & \underline{39.0} & 57.1 & \underline{27.4} & 28.6 & 12.5 \\
w/o Removal & 56.7 & \underline{26.0} & \underline{45.8} & 16.4 & \underline{56.9} & \underline{25.6} & \underline{71.4} & \textbf{40.5} & \textbf{58.3} & 27.0 & 50.0 & 22.0 \\
\midrule
\textbf{\HSGNav{} (Ours)} & \textbf{58.1} & \textbf{28.5} & \textbf{47.5} & \textbf{19.5} & \textbf{58.7} & \textbf{28.8} & \underline{71.4} & \textbf{40.5} & \textbf{58.3} & \textbf{28.3} & \textbf{57.1} & \textbf{29.3} \\
\bottomrule
\end{tabular*}
\end{table*}

\subsubsection{Evaluation Metrics}
To assess \textbf{navigation performance}, we report Success Rate (SR) for task completion and Success weighted by Path Length (SPL) for path efficiency.
To evaluate \textbf{hypothesis effectiveness}, we measure evidence retention rate (ERR), false-removal recovery (FRR), revisit rate (RR), and wasted travel ratio (WTR):
\begin{equation}
\label{eq:hypothesis-metrics}
\thickmuskip=3mu
\mathrm{ERR}=\tfrac{N_k}{N_d},\;
\mathrm{FRR}=\tfrac{N_r}{N_e},\;
\mathrm{RR}=\tfrac{N_v}{N_c},\;
\mathrm{WTR}=\tfrac{l_w}{l^a}.
\end{equation}
For ERR, we replay all three sequence layouts in each of the seven scenes (21 sequences), using their respective object placements; \HSG{} and Full Inst. receive identical RGB-D streams and detection proposals at three views per metre.
$N_d$ counts shared-detected target placement events: returns to earlier locations count separately, while repeated detections within an event count once.
$N_k$ counts events whose category remains in final memory linked to an image containing the corresponding GT-verified target from that event.
Both methods are assessed with the same retention criterion.
$N_e$ counts erroneous description invalidations; $N_r$ counts recoveries through new, retrievable target-matched descriptions from fresh same-anchor observations while the target remains there.
For RR/WTR, $N_c$ counts final rechecks after R3 and $N_v$ those selecting R3-rejected anchors with GT-confirmed continued target absence.
The corresponding travel distances, $l_w$ and $l^a$, measure travel toward those anchors and total recheck travel, excluding start-pose resets.
\textbf{Efficiency} is measured by VLM Call Count (VCC), VLM Image Count (VIC), and Average Path Length (APL), reported in metres.

\subsubsection{Implementation Details}
\label{sec:implementation-details}
Lifelong runs use Habitat-Sim with HM3D-Semantics-v0.2~\cite{habitat}, a 1\,m success threshold, and GOAT-Bench's subtask budget~\cite{goatbench}.
Baselines and ablations in Table~\ref{tab:dynamic_bench_mega} share the target vocabulary, GPT-4o (\texttt{gpt-4o-2024-08-06}), and 1280$\times$1280 RGB-D observations at 1.5\,m height with 120$^\circ$ horizontal FOV; VLM inputs are 360$\times$360.
They use ConceptGraph-style association, YOLO-World~\cite{yolo} with full-frame confidence $0.4$, and SAM~\cite{sam}.
+Det and Full Inst.\ share support-crop detections at confidence $0.1$: +Det retains the original add-only graphs and navigation, while Full Inst.\ instantiates proposals, with +Removal adding BDV-based instance removal.
OpenIN~\cite{openin} and IVAM-Nav~\cite{huang2026lisonav} are discussed qualitatively because their implementations were unavailable during evaluation.
Scene categories define the detector vocabulary; GPT-5.2 selects anchor categories using functional positive/negative examples, and matching detections become anchors.
Further details appear in the supplementary material.

\subsection{Lifelong Navigation Performance}
\label{sec:lifelong-results}

\paragraph{GOAT-Bench.}
Table~\ref{tab:goat_bench} evaluates static-scene lifelong navigation with category-name, natural-language description, and reference-image goals.
\HSGNav{} achieves 71.2\% SR and 50.3\% SPL, exceeding 3D-Mem by 2.1 and 1.4 percentage points.
The supplementary breakdown reports at least 67.0\% SR for each modality, suggesting that retrieval remains effective across different forms of goal evidence.

\paragraph{FAROS-Bench.}
On FAROS-Bench (Table~\ref{tab:dynamic_bench_mega}), \HSGNav{} achieves 58.1\% overall SR and 28.5\% SPL.
Its 25.7-point Small SR gain over the stronger baseline supports retaining surface-object evidence for small-target search.
Dynamic SR reaches 57.1\%, more than doubling both baselines, consistent with verification and memory revision mitigating stale-memory errors.
To isolate the effect of additional detections, +Det supplies identical surface-crop proposals while retaining each baseline's entity-instantiation pipeline and navigation.
MSGNav gains only 0.9 points in overall SR, while both baselines lose SPL.
These limited gains suggest that additional detections also require effective representation and updating.

\begin{table}[t]
\begingroup
\centering
\caption{\textbf{Efficiency analysis.} Mean VLM calls (VCC), input images (VIC), and path length (APL, m) per FAROS-Bench subtask.}
\label{tab:vlm_efficiency}
\aaaitabsetup
\begin{tabular*}{\linewidth}{@{}@{\extracolsep{\fill}}l@{\extracolsep{\fill}}c@{\extracolsep{\fill}}c@{\extracolsep{\fill}}c@{}}
\toprule
\textbf{Method} & \textbf{VCC} $\downarrow$ & \textbf{VIC} $\downarrow$ & \textbf{APL} $\downarrow$ \\
\midrule
MSGNav~\cite{msgnav} & 89.2 & 714.4 & 25.5 \\
3D-Mem~\cite{3dmem} & 36.8 & 657.3 & 22.7 \\
w/o SAMR            & \underline{21.6} & \underline{176.4} & 19.8 \\
w/o CGN             & 44.6 & 220.0 & \underline{18.5} \\
\midrule
\textbf{\HSGNav{} (Ours)} & \textbf{18.2} & \textbf{102.6} & \textbf{17.4} \\
\bottomrule
\end{tabular*}
\endgroup
\par\vspace{8pt}
\begingroup
\centering
\caption{\textbf{Relocation SR (\%).} In/Cross: within-/cross-anchor; \mbox{R1--R3}: repeated relocation; SR@1: fixed-target recheck.}
\label{tab:relocalization}
\footnotesize
\setlength{\tabcolsep}{1.5pt}
\renewcommand{\arraystretch}{1.12}
\begin{tabular*}{\linewidth}{@{}l@{\extracolsep{\fill}}cccccc@{}}
\toprule
\multirow{2}{*}{\textbf{Method}} & \multicolumn{2}{c}{Single-task} & \multicolumn{3}{c}{Continuous} & Recheck \\
\cmidrule(lr){2-3}\cmidrule(lr){4-6}\cmidrule(lr){7-7}
& In & Cross & R1 & R2 & R3 & SR@1 \\
\midrule
DualMap~\cite{dualmap} & 64.8 & 60.3 & 60.8 & 54.9 & 49.0 & 47.1 \\
OASIS~\cite{liao2026oasis} & \underline{75.9} & \underline{73.6} & 74.5 & 68.6 & 64.7 & 62.7 \\
\midrule
\HSGNav{} w/o Removal & --- & --- & \textbf{82.4} & 70.6 & 60.8 & 52.9 \\
\textbf{\HSGNav{} (Ours)} & \textbf{85.2} & \textbf{83.0} & \textbf{82.4} & \textbf{80.4} & \textbf{78.4} & \textbf{76.5} \\
\bottomrule
\end{tabular*}
\endgroup
\end{table}

\begin{figure}[t]
\centering
\includegraphics[width=\linewidth]{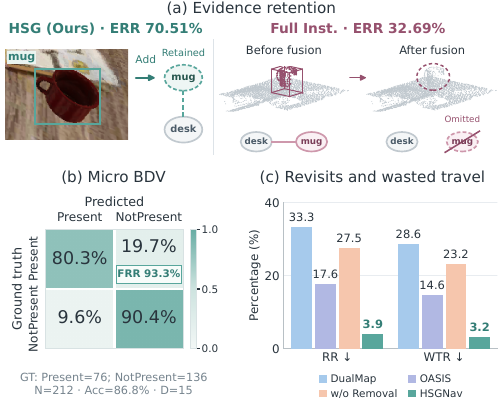}
\caption{\textbf{Hypothesis effectiveness.}
\textbf{(a)}~Evidence retention (ERR, $\uparrow$) and information omission.
\textbf{(b)}~Micro BDV confusion matrix and recovery (FRR); $D$ counts erroneous invalidations.
\textbf{(c)}~Recheck revisit rate (RR) and wasted travel ratio (WTR).}
\label{fig:hypothesis_analysis}
\end{figure}

\subsection{Ablation Studies}
\label{sec:ablations}

\paragraph{Representation and Update Controls.}
Representation controls share OCE, BDV, CGN, and the VLM backend to isolate memory representation effects (Table~\ref{tab:dynamic_bench_mega}).
``Traditional'' and ``Multi-modal'' use ConceptGraph-style textual~\cite{conceptgraphs} and MSGNav-style visual graphs~\cite{msgnav}.
``Anchor-Centric'' groups mapped instances and views around stable anchors~\cite{liao2026oasis}; ``Entity-Level Only'' omits hypotheses.
\HSGNav{} improves Tiny/Small SR over Anchor-Centric by 20.4/26.6 points, versus 2.4 on Large targets, suggesting that separate surface hypotheses mainly benefit small-object search.
``Full Inst.'' instantiates identical crop detections as entities; ``Full Inst.\ + Removal'' removes rejected instances and invalidates linked retrieval evidence.
With removal, Small SR is 35.8\% versus 58.7\% for \HSGNav{}.
This navigation comparison is complemented by the matched-detection evidence-retention analysis in Fig.~\ref{fig:hypothesis_analysis}(a).

\paragraph{Module Ablations.}
We ablate modules with full \HSG{} construction (Table~\ref{tab:dynamic_bench_mega}).
Without SAMR, scale-agnostic retrieval jointly inspects macro/micro frames from at most three anchor-linked groups; SR/SPL fall to 52.4\%/23.9\%, supporting scale-aware evidence selection.
Frontier-only exploration without OCE lowers SR by 10.0 points, especially for tiny/small targets, suggesting that surface observation complements spatial coverage.
Removing CGN reduces SPL from 28.5\% to 25.4\% with little SR change, supporting goal commitment for path efficiency.
Without BDV, Dynamic SR falls from 57.1\% to 28.6\%; retaining verification but disabling Removal yields 50.0\% while preserving Static SR.
These results indicate complementary roles for target verification and persistent invalidation of rejected evidence in subsequent dynamic search.

\paragraph{Efficiency Analysis.}
\HSGNav{} averages 18.2 VLM calls and 102.6 images per subtask, approximately half the calls and one-sixth the images of 3D-Mem (Table~\ref{tab:vlm_efficiency}).
Removing CGN raises calls to 44.6 through transit replanning; removing SAMR raises image inputs to 176.4 through joint macro/micro inspection.
These increases link the savings to goal commitment and scale-aware evidence selection.
The lowest APL (17.4\,m) also indicates that reduced model usage does not require longer navigation paths.

\subsection{Revisable Hypothesis Memory}
\label{sec:relocalization}

\paragraph{Relocation Robustness.}
Table~\ref{tab:relocalization} evaluates relocation robustness in pre-mapped scenes under matched observations, controllers, and search budgets.
\HSGNav{} outperforms OASIS in single-task In-/Cross-anchor search and degrades less under repeated relocation.
Without Removal, verification and new evidence admission remain active, but rejected evidence is retained.
Both versions match at R1, yet R3 SR falls to 60.8\% without Removal versus 78.4\% for the full system, with poorer fixed-target rechecks.
On the fixed-target recheck, SR is 76.5\% versus 52.9\%, showing that stale records can mislead retrieval even without another relocation.
The growing gap suggests that fresh observations and verification cannot fully offset stale evidence; persistent invalidation supports retrieval across repeated changes.

\paragraph{Hypothesis Effectiveness.}
Figure~\ref{fig:hypothesis_analysis}(a) shows how Full Inst. can fuse a mug into furniture, losing object-specific retrieval evidence.
\HSG{} preserves anchor-linked labels and images, achieving 70.51\% ERR versus 32.69\% for Full Inst. on shared-detected placement events.
The matched detections control for detector recall; ERR measures retained visual evidence without requiring an independent 3D target instance.
Micro BDV achieves 86.8\% accuracy over 212 checks (Fig.~\ref{fig:hypothesis_analysis}(b)).
After erroneous invalidation, fresh observations recover retrievable target-matched descriptions in 93.3\% of cases, allowing current evidence to support renewed retrieval while rejected historical records remain excluded.
Compared with w/o Removal, RR falls from 27.5\% to 3.9\% and WTR from 23.2\% to 3.2\% (Fig.~\ref{fig:hypothesis_analysis}(c)), linking persistent invalidation to fewer stale-anchor choices and less wasted travel.

\section{Conclusion}
We presented \HSGNav{}, which anchors revisable surface-object hypotheses to persistent furniture entities, retaining small-object evidence without independent 3D reconstruction.
Scale-adaptive retrieval, surface exploration, and verification support evidence acquisition, reuse, and revision.
Extensive experiments demonstrate performance gains, highlighting the effectiveness of revisable hypothesis memory in preserving evidence and supporting multi-scale lifelong navigation under object relocation.

\paragraph{Limitations and Future Work.}
Our method uses persistent furniture anchors as spatial references for surface hypotheses throughout continuous navigation. A natural extension is to maintain and reassociate these hypotheses as support objects move, enabling memory revision to follow changes in both surface objects and their anchors.

\section*{LLM Usage Statement}
GPT-5.2 was used to identify semantic anchor categories from scene category lists and to generate FAROS-Bench episode scripts.
Generated object placements were subsequently verified and curated by the authors in simulation.
GPT-4o was used as the vision-language backend for navigation, and Qwen2.5-VL-7B was evaluated as an alternative backend, as detailed in the paper and supplementary material.
ChatGPT and Codex were used to assist with manuscript rewriting and code development, and ChatGPT was also used to generate the object icons (e.g., apples) in Figures~1--3.
The authors take full responsibility for the submitted work.

{\small\bibliographystyle{ieeenat_fullname}\bibliography{references}}

@inproceedings{clip,
  title={Learning transferable visual models from natural language supervision},
  author={Radford, Alec and Kim, Jong Wook and Hallacy, Chris and Ramesh, Aditya and Goh, Gabriel and Agarwal, Sandhini and Sastry, Girish and Askell, Amanda and Mishkin, Pamela and Clark, Jack and others},
  booktitle={International conference on machine learning},
  pages={8748--8763},
  year={2021},
  organization={PmLR}
}

@inproceedings{blip,
  title={Blip: Bootstrapping language-image pre-training for unified vision-language understanding and generation},
  author={Li, Junnan and Li, Dongxu and Xiong, Caiming and Hoi, Steven},
  booktitle={International conference on machine learning},
  pages={12888--12900},
  year={2022},
  organization={PMLR}
}

@inproceedings{seem,
 author = {Zou, Xueyan and Yang, Jianwei and Zhang, Hao and Li, Feng and Li, Linjie and Wang, Jianfeng and Wang, Lijuan and Gao, Jianfeng and Lee, Yong Jae},
 booktitle = {Advances in Neural Information Processing Systems},
 editor = {A. Oh and T. Naumann and A. Globerson and K. Saenko and M. Hardt and S. Levine},
 pages = {19769--19782},
 publisher = {Curran Associates, Inc.},
 title = {Segment Everything Everywhere All at Once},
 url = {https://proceedings.neurips.cc/paper_files/paper/2023/file/3ef61f7e4afacf9a2c5b71c726172b86-Paper-Conference.pdf},
 volume = {36},
 year = {2023}
}

@inproceedings{openscene,
      title     = {OpenScene: 3D Scene Understanding with Open Vocabularies},
      author    = {Peng, Songyou and Genova, Kyle and Jiang, Chiyu "Max" and Tagliasacchi, Andrea and Pollefeys, Marc and Funkhouser, Thomas},
      booktitle = {Proceedings of the IEEE/CVF Conference on Computer Vision and Pattern Recognition (CVPR)},
      year      = {2023}
  }

@inproceedings{CLIP-Fields, series={RSS2023},
   title={CLIP-Fields: Weakly Supervised Semantic Fields for Robotic Memory},
   url={http://dx.doi.org/10.15607/RSS.2023.XIX.074},
   DOI={10.15607/rss.2023.xix.074},
   booktitle={Robotics: Science and Systems XIX},
   publisher={Robotics: Science and Systems Foundation},
   author={(Mahi)Shafiullah, Nur and Paxton, Chris and Pinto, Lerrel and Chintala, Soumith and Szlam, Arthur},
   year={2023},
   month={July}, collection={RSS2023} }

@inproceedings{Open-fusion,
  title={Open-fusion: Real-time open-vocabulary 3d mapping and queryable scene representation},
  author={Yamazaki, Kashu and Hanyu, Taisei and Vo, Khoa and Pham, Thang and Tran, Minh and Doretto, Gianfranco and Nguyen, Anh and Le, Ngan},
  booktitle={2024 IEEE International Conference on Robotics and Automation (ICRA)},
  pages={9411--9417},
  year={2024},
  organization={IEEE}
}

@inproceedings{CLIP-FO3D,
  title     = {{CLIP-FO3D: Learning Free Open-World 3D Scene Representations from 2D Dense CLIP}},
  author    = {Zhang, Junbo and Dong, Runpei and Ma, Kaisheng},
  booktitle = {IEEE/CVF International Conference on Computer Vision Workshops},
  year      = {2023},
  pages     = {2040-2051},
  doi       = {10.1109/ICCVW60793.2023.00219},
  url       = {https://mlanthology.org/iccvw/2023/zhang2023iccvw-clipfo3d/}
}

@inproceedings{3D-Concept-Grounding,
 author = {Hong, Yining and Du, Yilun and Lin, Chunru and Tenenbaum, Josh and Gan, Chuang},
 booktitle = {Advances in Neural Information Processing Systems},
 editor = {S. Koyejo and S. Mohamed and A. Agarwal and D. Belgrave and K. Cho and A. Oh},
 pages = {7769--7782},
 publisher = {Curran Associates, Inc.},
 title = {3D Concept Grounding on Neural Fields},
 url = {https://proceedings.neurips.cc/paper_files/paper/2022/file/330073c95529dae593936d387edac58c-Paper-Conference.pdf},
 volume = {35},
 year = {2022}
}

@inproceedings{hydra,
    title={Hydra: A Real-time Spatial Perception System for {3D} Scene Graph Construction and Optimization},
    fullauthor={Nathan Hughes, Yun Chang, and Luca Carlone},
    author={N. Hughes and Y. Chang and L. Carlone},
    booktitle={Robotics: Science and Systems (RSS)},
    year={2022},
}

@article{Kimera,
  title={Kimera: From SLAM to spatial perception with 3D dynamic scene graphs},
  author={Antoni Rosinol and Andrew Violette and Marcus Abate and Nathan Hughes and Yun Chang and J. Shi and Arjun Gupta and Luca Carlone},
  journal={The International Journal of Robotics Research},
  year={2021},
  volume={40},
  pages={1510 - 1546},
  url={https://api.semanticscholar.org/CorpusID:231632715}
}

@InProceedings{SceneGraphFusion,
    author    = {Wu, Shun-Cheng and Wald, Johanna and Tateno, Keisuke and Navab, Nassir and Tombari, Federico},
    title     = {SceneGraphFusion: Incremental 3D Scene Graph Prediction From RGB-D Sequences},
    booktitle = {Proceedings of the IEEE/CVF Conference on Computer Vision and Pattern Recognition (CVPR)},
    month     = {June},
    year      = {2021},
    pages     = {7515-7525}
}

@INPROCEEDINGS{3D-Scene-Graph,
  author={Armeni, Iro and He, Zhi-Yang and Zamir, Amir and Gwak, Junyoung and Malik, Jitendra and Fischer, Martin and Savarese, Silvio},
  booktitle={2019 IEEE/CVF International Conference on Computer Vision (ICCV)}, 
  title={3D Scene Graph: A Structure for Unified Semantics, 3D Space, and Camera}, 
  year={2019},
  volume={},
  number={},
  pages={5663-5672},
  doi={10.1109/ICCV.2019.00576}}

@inproceedings{VGfM,
  title={Visual Graphs from Motion (VGfM): Scene understanding with object geometry reasoning},
  author={Gay, Paul and James, Stuart and Del Bue, Alessio},
  booktitle={Asian conference on computer vision (ACCV)},
  year={2018}
}

@inproceedings{conceptgraphs,
  title={Conceptgraphs: Open-vocabulary 3d scene graphs for perception and planning},
  author={Gu, Qiao and Kuwajerwala, Ali and Morin, Sacha and Jatavallabhula, Krishna Murthy and Sen, Bipasha and Agarwal, Aditya and Rivera, Corban and Paul, William and Ellis, Kirsty and Chellappa, Rama and others},
  booktitle={2024 IEEE International Conference on Robotics and Automation (ICRA)},
  pages={5021--5028},
  year={2024},
  organization={IEEE}
}

@article{3D-Semantic-Scene-Graph,
   title={3-D Scene Graph: A Sparse and Semantic Representation of Physical Environments for Intelligent Agents},
   volume={50},
   ISSN={2168-2275},
   url={http://dx.doi.org/10.1109/TCYB.2019.2931042},
   DOI={10.1109/tcyb.2019.2931042},
   number={12},
   journal={IEEE Transactions on Cybernetics},
   publisher={Institute of Electrical and Electronics Engineers (IEEE)},
   author={Kim, Ue-Hwan and Park, Jin-Man and Song, Taek-jin and Kim, Jong-Hwan},
   year={2020},
   month=Dec, pages={4921–4933} }

@inproceedings{goatbench,
      title={GOAT-Bench: A Benchmark for Multi-Modal Lifelong Navigation}, 
      author={Mukul Khanna* and Ram Ramrakhya* and Gunjan Chhablani and Sriram Yenamandra and Theophile Gervet and Matthew Chang and Zsolt Kira and Devendra Singh Chaplot and Dhruv Batra and Roozbeh Mottaghi},
      year={2024},
      booktitle={CVPR},
}

@inproceedings{hm3d-ovon,
  title={Hm3d-ovon: A dataset and benchmark for open-vocabulary object goal navigation},
  author={Yokoyama, Naoki and Ramrakhya, Ram and Das, Abhishek and Batra, Dhruv and Ha, Sehoon},
  booktitle={2024 IEEE/RSJ International Conference on Intelligent Robots and Systems (IROS)},
  pages={5543--5550},
  year={2024},
  organization={IEEE}
}

@inproceedings{mtu3d,
  title={Move to understand a 3d scene: Bridging visual grounding and exploration for efficient and versatile embodied navigation},
  author={Zhu, Ziyu and Wang, Xilin and Li, Yixuan and Zhang, Zhuofan and Ma, Xiaojian and Chen, Yixin and Jia, Baoxiong and Liang, Wei and Yu, Qian and Deng, Zhidong and others},
  booktitle={Proceedings of the IEEE/CVF International Conference on Computer Vision},
  pages={8120--8132},
  year={2025}
}

@inproceedings{cows,
  title={Cows on pasture: Baselines and benchmarks for language-driven zero-shot object navigation},
  author={Gadre, Samir Yitzhak and Wortsman, Mitchell and Ilharco, Gabriel and Schmidt, Ludwig and Song, Shuran},
  booktitle={Proceedings of the IEEE/CVF Conference on Computer Vision and Pattern Recognition},
  pages={23171--23181},
  year={2023}
}

@inproceedings{esc,
  title={Esc: Exploration with soft commonsense constraints for zero-shot object navigation},
  author={Zhou, Kaiwen and Zheng, Kaizhi and Pryor, Connor and Shen, Yilin and Jin, Hongxia and Getoor, Lise and Wang, Xin Eric},
  booktitle={International Conference on Machine Learning},
  pages={42829--42842},
  year={2023},
  organization={PMLR}
}

@inproceedings{openfmnav,
  title={Openfmnav: Towards open-set zero-shot object navigation via vision-language foundation models},
  author={Kuang, Yuxuan and Lin, Hai and Jiang, Meng},
  booktitle={Findings of the Association for Computational Linguistics: NAACL 2024},
  pages={338--351},
  year={2024}
}

@inproceedings{vlfm,
  title={Vlfm: Vision-language frontier maps for zero-shot semantic navigation},
  author={Yokoyama, Naoki and Ha, Sehoon and Batra, Dhruv and Wang, Jiuguang and Bucher, Bernadette},
  booktitle={2024 IEEE International Conference on Robotics and Automation (ICRA)},
  pages={42--48},
  year={2024},
  organization={IEEE}
}

@article{dualmap,
  title={DualMap: Online Open-Vocabulary Semantic Mapping for Natural Language Navigation in Dynamic Changing Scenes},
  author={Jiang, Jiajun and Zhu, Yiming and Wu, Zirui and Song, Jie},
  journal={IEEE Robotics and Automation Letters},
  year={2025},
  publisher={IEEE}
}

@inproceedings{liao2026oasis,
  title     = {{OASIS}: Online Anchor-centric Scene Inference System for Open-Vocabulary Retrieval and Navigation in Dynamic Environments},
  author    = {Liao, Kaiyao and Li, Qingfeng and Chen, Chen and Wu, Xiaoze and Sun, Qing and Niu, Jianwei},
  booktitle = {Proceedings of the 34th ACM International Conference on Multimedia},
  year      = {2026},
  note      = {To appear}
}

@inproceedings{hovsg,
  title={Hierarchical open-vocabulary 3d scene graphs for language-grounded robot navigation},
  author={Werby, Abdelrhman and Huang, Chenguang and B{\"u}chner, Martin and Valada, Abhinav and Burgard, Wolfram},
  booktitle={First Workshop on Vision-Language Models for Navigation and Manipulation at ICRA 2024},
  year={2024}
}

@article{sgnav,
  title={Sg-nav: Online 3d scene graph prompting for llm-based zero-shot object navigation},
  author={Yin, Hang and Xu, Xiuwei and Wu, Zhenyu and Zhou, Jie and Lu, Jiwen},
  journal={Advances in neural information processing systems},
  volume={37},
  pages={5285--5307},
  year={2024}
}

@inproceedings{unigoal,
  title={Unigoal: Towards universal zero-shot goal-oriented navigation},
  author={Yin, Hang and Xu, Xiuwei and Zhao, Linqing and Wang, Ziwei and Zhou, Jie and Lu, Jiwen},
  booktitle={Proceedings of the IEEE/CVF Conference on Computer Vision and Pattern Recognition},
  pages={19057--19066},
  year={2025}
}

@inproceedings{3dmem,
  title={3D-mem: 3D scene memory for embodied exploration and reasoning},
  author={Yang, Yuncong and Yang, Han and Zhou, Jiachen and Chen, Peihao and Zhang, Hongxin and Du, Yilun and Gan, Chuang},
  booktitle={Proceedings of the Computer Vision and Pattern Recognition Conference},
  pages={17294--17303},
  year={2025}
}

@inproceedings{msgnav,
  title={Msgnav: Unleashing the power of multi-modal 3d scene graph for zero-shot embodied navigation},
  author={Huang, Xun and Zhao, Shijia and Wang, Yunxiang and Lu, Xin and Zhang, Wanfa and Qu, Rongsheng and Li, Weixin and Wang, Yunhong and Wen, Chenglu},
  booktitle={Proceedings of the IEEE/CVF Conference on Computer Vision and Pattern Recognition},
  pages={37154--37163},
  year={2026}
}

@inproceedings{idu,
  title={Interaction-Driven Updates: 3D Scene Graph Maintenance During Robot Task Execution},
  author={Li, Qingfeng and Zhang, Xinlei and Chen, Chen and Zhao, Haochen and Niu, Jianwei},
  booktitle={2025 IEEE International Conference on Robotics and Automation (ICRA)},
  pages={11933--11939},
  year={2025},
  organization={IEEE}
}

@article{openin,
  title={Openin: Open-vocabulary instance-oriented navigation in dynamic domestic environments},
  author={Tang, Yujie and Wang, Meiling and Deng, Yinan and Zheng, Zibo and Deng, Jingchuan and Zuo, Sibo and Yue, Yufeng},
  journal={IEEE Robotics and Automation Letters},
  volume={10},
  number={9},
  pages={9256--9263},
  year={2025},
  publisher={IEEE}
}

@inproceedings{vlmnav,
      title={End-to-End Navigation with VLMs: Transforming Spatial Reasoning into Question-Answering},
      author={Dylan Goetting and Himanshu Gaurav Singh and Antonio Loquercio},
      booktitle={Workshop on Language and Robot Learning: Language as an Interface},
      year={2024},
    }

@misc{dynavlm,
      title={DyNaVLM: Zero-Shot Vision-Language Navigation System with Dynamic Viewpoints and Self-Refining Graph Memory}, 
      author={Zihe Ji and Huangxuan Lin and Yue Gao},
      year={2025},
      eprint={2506.15096},
      archivePrefix={arXiv},
      primaryClass={cs.RO},
      url={https://arxiv.org/abs/2506.15096}, 
}

@inproceedings{tango,
  title={Tango: training-free embodied ai agents for open-world tasks},
  author={Ziliotto, Filippo and Campari, Tommaso and Serafini, Luciano and Ballan, Lamberto},
  booktitle={Proceedings of the Computer Vision and Pattern Recognition Conference},
  pages={24603--24613},
  year={2025}
}

@article{habitat, 
title={Habitat-Matterport 3D Semantics Dataset}, author={Yadav, Karmesh and Ramrakhya, Ram and Ramakrishnan, Santhosh Kumar and Gervet, Theo and Turner, John and Gokaslan, Aaron and Maestre, Noah and Chang, Angel Xuan and Batra, Dhruv and Savva, Manolis and others}, journal={arXiv preprint arXiv:2210.05633}, year={2022}, url={https://arxiv.org/abs/2210.05633} }

@inproceedings{yolo,
  title={YOLO-World: Real-Time Open-Vocabulary Object Detection},
  author={Cheng, Tianheng and Song, Lin and Ge, Yixiao and Liu, Wenyu and Wang, Xinggang and Shan, Ying},
  booktitle={Proc. IEEE Conf. Computer Vision and Pattern Recognition (CVPR)},
  year={2024}
}

@article{sam,
  title={Segment Anything},
  author={Kirillov, Alexander and Mintun, Eric and Ravi, Nikhila and Mao, Hanzi and Rolland, Chloe and Gustafson, Laura and Xiao, Tete and Whitehead, Spencer and Berg, Alexander C. and Lo, Wan-Yen and Doll{\'a}r, Piotr and Girshick, Ross},
  journal={arXiv:2304.02643},
  year={2023}
}

@inproceedings{cmar,
   title={CMAR-search: Commonsense and Memory Augmented Reasoning for Object Search in Dynamic Interactive Environments},
   url={http://dx.doi.org/10.1109/ICRA57385.2026.11697035},
   DOI={10.1109/icra57385.2026.11697035},
   booktitle={2026 IEEE International Conference on Robotics and Automation (ICRA)},
   publisher={IEEE},
   author={Liao, Kaiyao and Li, Qingfeng and Zhang, Xinlei and Chen, Chen and Sun, Qing and Niu, Jianwei},
   year={2026},
   month=June, pages={18586–18593} }

@misc{huang2026lisonav,
  title={Lifelong Small-Object Navigation in Changing Object Layouts: A Benchmark and Method},
  author={Huang, Jiagan and Zhou, Zikun and Ni, Zijian and Wang, Hongpeng and Lu, Guangming and Yu, Jun and Pei, Wenjie},
  year={2026},
  eprint={2610.10125},
  archivePrefix={arXiv},
  primaryClass={cs.RO},
  url={https://arxiv.org/abs/2610.10125}
}
\end{document}